\documentclass[preprint]{article}
\usepackage{tmlr}

\usepackage{amsmath,amsfonts,bm}

\def\eqref#1{equation~\ref{#1}}

\def\1{\bm{1}}

\DeclareMathAlphabet{\mathsfit}{\encodingdefault}{\sfdefault}{m}{sl}
\SetMathAlphabet{\mathsfit}{bold}{\encodingdefault}{\sfdefault}{bx}{n}

\usepackage{hyperref}
\usepackage{url}
\usepackage{graphicx}
\usepackage{algorithm}
\usepackage{algorithmic}
\usepackage{amsmath}
\usepackage{bm}
\usepackage{latexsym}
\usepackage{booktabs}
\usepackage{wrapfig}

\title{CluMo: Cluster-based Modality Fusion Prompt for Continual Learning in Visual Question Answering}

\author{\name Yuliang Cai  \email caiyulia@usc.edu\\
      \addr 
      University of Southern California
      \\
      \name Mohammad Rostami \email  rostamim@usc.edu \\
     \addr  
      University of Southern California}

\def\month{MM}  
\def\year{YYYY} 
\def\openreview{\url{https://openreview.net/forum?id=XXXX}} 

\begin{document}

\maketitle

\begin{abstract}
Large  vision-language models (VLMs) have shown significant performance boost in various application domains. However, adopting them to deal with several sequentially encountered tasks has been challenging because finetuning a VLM on a task normally leads to reducing its generalization power and the capacity of learning new tasks as well as causing catastrophic forgetting on previously learned tasks. Enabling using VLMs in multimodal continual learning (CL) settings can help to address such scenarios. To improve generalization capacity and prevent catastrophic forgetting, we propose a novel prompt-based CL method for VLMs, namely $\textbf{Clu}$ster-based $\textbf{Mo}$dality Fusion Prompt (\textbf{CluMo}). We design a novel \textbf{Key-Key-Prompt} pair, where each prompt is associated with a visual prompt key and a textual prompt key. We adopt a two-stage training strategy. During the first stage, the single-modal keys are trained via  $K$-means clustering algorithm to help select the best semantically matched prompt. During the second stage, the prompt keys are frozen, the selected prompt is attached to the input for training the VLM in the CL scenario. Experiments on two benchmarks demonstrate that our method achieves SOTA performance.
\end{abstract}

\section{Introduction}
Visual Question Answering (VQA) is a complicated  task, where the goal is  to answer questions described in natural language (text) about  a given input image. Addressing VQA requires understanding and fusion of information from both the visual and textual domains to generate accurate responses. Recently,  significant advancements in addressing VQA tasks have emerged due to the development of pre-trained large vision-language models (VLMs) \cite{radford2021learning, kim2021vilt}. Despite these advances, one of the persistent challenges in VQA tasks is the ability to adapt a VLM in CL setting to avoid finetuning a copy of an underlying VLM per task. In a CL setting, we learn new tasks and aim for continuously improving the model performance without forgetting previously learned knowledge, also known as catastrophic forgetting \cite{french1999catastrophic}.
To address  catastrophic forgetting, a group of CL algorithms are deployed. Regularization-based methods \cite{Kirkpatrick_2017, jin2021learn}  constrain the drastic parameter shift when learning  new tasks. Expansion-based methods \cite{douillard2022dytox,cai2023taskattentive} expand the model with small portion of additional weights and use the expanded weights to learn the new incoming tasks. Rehearsal-based methods \cite{rebuffi2017icarl,rostami2021lifelong} store a representative subset of the training dataset for each task into a small memory buffer and replay them back during the learning of the current task to maintain the encoded knowledge of the previously learned tasks. More recently, prompt-based methods \cite{wang2022learning,wang2022dualprompt,cai2023task} aim to use prompts that contains task-specific or semantic-specific information to prevent catastrophic forgetting.  A prompt is  attached to the embedded features of the input to adapt the model to focus on the specific characteristics of the input task that has been learned before.

\begin{figure}[]
\begin{center}
\includegraphics[scale=0.5]{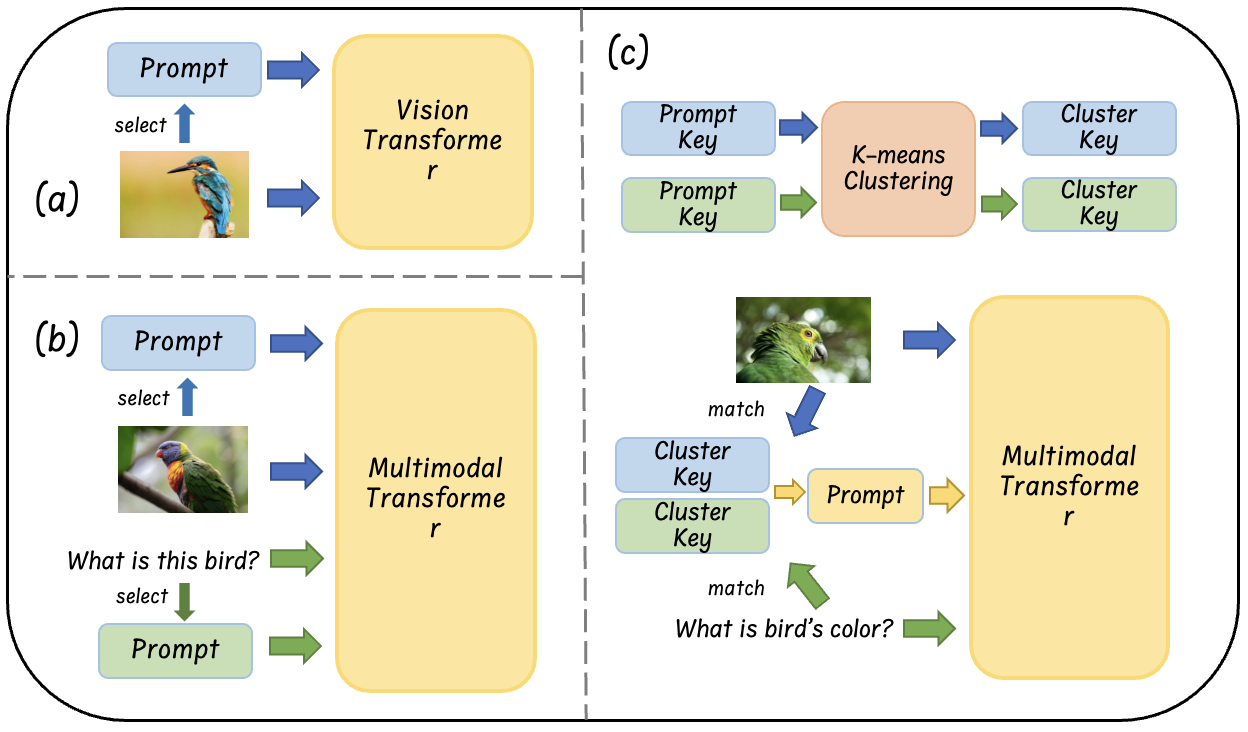}
\end{center}
\caption{Comparison between existing prompt-based CL methods and our proposed method: $\textbf{(a)}$ Uni-modal based methods   use image feature to select prompts from a prompt pool. $\textbf{(b)}$ Multi-modal based methods use the image features to select image prompts and use the text features to select the text prompts. $\textbf{(c)}$ We first train the prompt key using a clustering algorithm to form a cluster key and use the combination of the cluster key from both modalities to select the fusion prompt.}
\label{fig:comp}
\end{figure}

Most existing CL methods consider unimodal, i.e., vision-only or language-only, settings and hence are inapplicable to address VQA tasks.
To tackle this shortcoming, we propose a novel two-stage prompt learning-based CL method, namely cluster-based modality fusion prompt (CluMo). Figure \ref{fig:comp} visualizes the high-level idea of our approach. Our method adopts a pre-trained VLM as its backbone  and benefits from a clustering-based modal-specific key strategy to boost generalization capacity and minimize catastrophic forgetting. More specifically,  we use a clustering-based algorithm to train visual-prompt keys and textual-prompt keys during the first stage. During the second stage, we assign each input image-question pair  with well-trained prompt keys to its corresponding visual keys and textual keys. We then use the combination of two modal-specific keys to find the best-matched prompt to adapt the model for the input task. We also benefit from knowledge distillation   during   training to further improve the performance. Our proposed method outperforms existing alternative methods. Our specific contribution includes:
\begin{itemize}
    \item  We propose a novel clustering-based prompt learning method for training VLMs in CL settings to address VQA tasks with vision-and-language inputs.
    \item  We use a two-stage training strategy to train the prompt keys before training the whole model to guarantee  optimal prompt selection that is necessary for generalization on the input.
    \item  We offer extensive experiments to demonstrate that the proposed approach achieves SOTA performance against CL existing methods and offer insight about the reason of this improved performance.
\end{itemize}

\section{Related Works}

\paragraph{Visual Question Answering}
Visual Question Answering (VQA) has been a pivotal task at the intersection of computer vision and natural language processing which led to advances on more complex tasks. Initially, VQA was formulated as a classification task in which answers are selected from a predefined set of answers \cite{agrawal2016vqa} and was solved by using CNNs for image feature extraction and RNNs for text processing. These models were too simple to be used in most practical cases. With the development of transformer and BERT-like models \cite{lu2019vilbert, li2019visualbert},   performance in VQA tasks has significantly been improved due to the better capacity of capturing the intricate relationship between two modalities and generating the response in the form of meaningful and descriptive texts. Despite these advances, VQA tasks are mostly studied in static settings \cite{goyal2017making,johnson2016clevr,marino2019okvqa}, where it is assumed that there is a single input task. As a result, most existing methods are inapplicable  in dynamic environments and settings such as continual learning (CL)~\cite{srinivasan2022climb} when we encounter several tasks over time. 

\paragraph{Prompt-Based Learning}
Prompt learning is a powerful technique for leveraging pre-trained language models to frame downstream tasks in NLP. It is more memory-efficient than  using Adapters \cite{pfeiffer2021adapterfusion} or LoRA \cite{hu2021lora} and has been used successfully to  guide  responses of VLMs for a particular task. 
The reason is that  additional prompt parameters are  in the form of small embeddings   which are directly added to the input to make the model task. 
As a result, the prompts are often much smaller in size compared to the layers of the model, leading to a minimal increase in the total number of parameters compared to using adapters or LoRA. Moreover, prompts can be stored in a memory-efficient way and retrieved dynamically based on the observed task in the input. This means that even when the model handles many tasks, the memory overhead remains low.
Browon et al.~\citeyear{brown2020language}   introduced the concept of prompt for the natural language instruction task to guide the model towards desired outputs. Prompt learning is based on providing a fixed function to condition a model so that it gets extra information token which specializes it to perform the down-stream task. Prompts are mostly considered as trainable parameters, either task-specific or domain-specific, to guide the model by obtaining task-specific knowledge \cite{lester2021power,li2021prefixtuning}. Prompt learning has also been found helpful in handling a single VQA task~\cite{hu2023promptcap}.

\paragraph{Prompt Learning for Continual Learning}
Prompt learning has been used in CL to prevent catastrophic forgetting when a large pre-trained models is trained on a stream of sequentially encountered tasks. It allows a single model to quickly adapt to new tasks in the stream without needing extensive retraining.  By using task-specific prompts, the model can retain and recall knowledge from earlier tasks, mitigating the issue of catastrophic forgetting. It also allows scalability to learning a large number of tasks since each task primarily requires learning or generating new prompts rather than retraining the entire model.
L2P \cite{wang2022learning} pioneered to connect prompt-based learning and CL. Instead of having a single shared prompt to learn all  tasks, L2P introduced the concept of ``prompt pool''  to maintain prompts for different tasks independently from each other. DualPrompt \cite{wang2022dualprompt} extended the idea of prompt pool in l2p by introducing  E-prompt and G-prompt. While E-prompt is task-specific, G-prompt encodes the knowledge used for all tasks to further allow knowledge sharing and transferring while mitigating negative transfer. S-Prompt \cite{wang2023sprompts}  applied clustering   to build the prompt pool with domain-specific prompts. These prompt learning methods for CL only consider single-modality, i.e., vision-only or text-only, and hence are sub-optimal for tasks with  multi-modal inputs such VQA when the modalities are related. Our method  benefits from the specific properties of multi-modal data to address VQA in CL settings using prompt learning and leads to performance improvements against these methods.

\begin{figure*}[]
\centering
\includegraphics[scale=0.54]{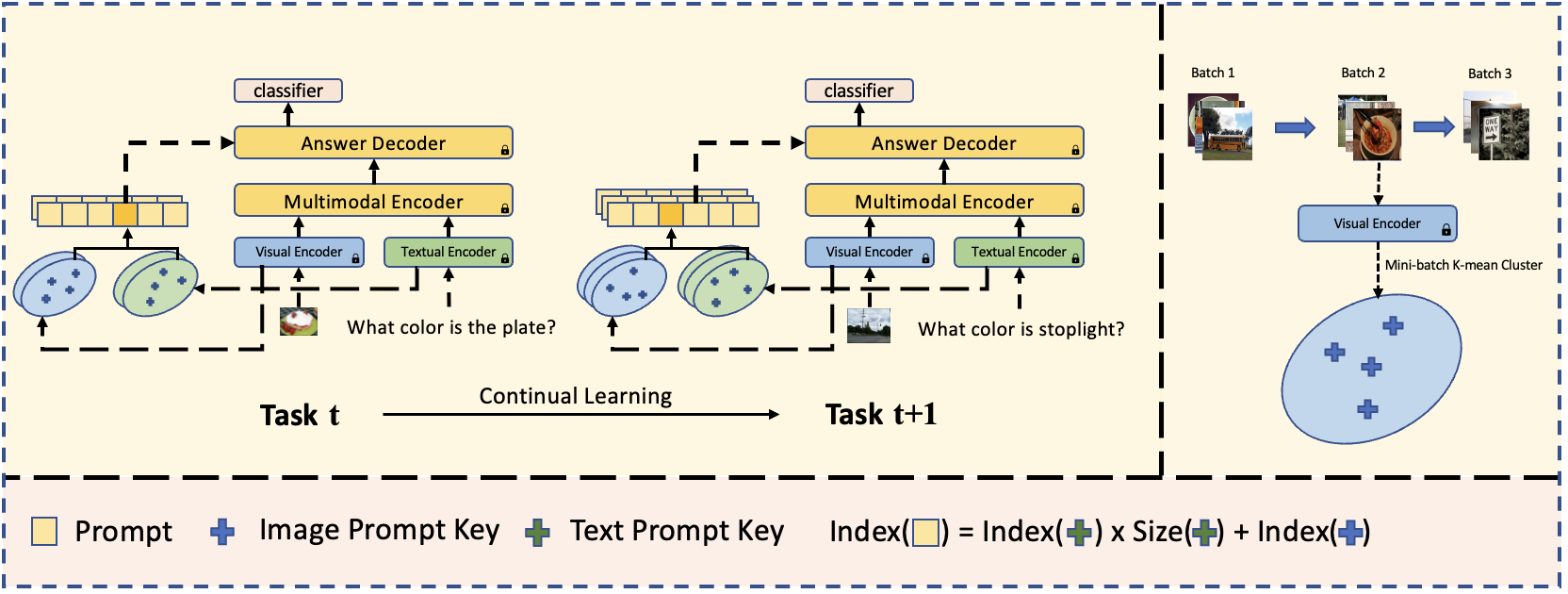}
\caption{Block diagram of the proposed approach: $\textbf{Left}$:  the backbone contains a pre-trained frozen visual encoder, a textual encoder, and a multimodal encoder. The answer decoder shares the same architecture as multimodal encoder. During the training phase, the fixed number of visual prompt keys, textual prompt keys, and a prompt pool will be added for each new task. $\textbf{Right}$: the procedure of visual prompt key training consists of training the modal-specific prompt key by a sequence of randomly selected batches of training data from current task until convergence is reached. Same procedure for textual prompt key.}
\label{fig:Big_Diagram1}
\end{figure*}

\section{Problem Description}
Consider a set of VQA tasks, $\{\mathcal{T}_i\}_{i=1}^T$, which are encountered  sequentially and each of them is from different domain. For each of the tasks,  a labeled training dataset $\mathcal{D}^i$ = $\{\langle(\bm{I}_i^j,\bm{L}_i^j)^i, y^j_i\rangle_{j=1}^{N_i}\}$ is accessible, $N_i$ denotes the size of dataset, $\bm{I}_i^j \in \mathbb{R}^{H\times W\times C}$ denotes the input image,   $\bm{L}_i^j \in \mathbb{R}^{L\times |V|}$ denotes the input text, and  $y^j_i$ denotes the text-typed discrete label. The order in which the VQA tasks are observed is not known in advance and the training data  points are assumed to be drawn iid from a task-specific joint distribution $p_i^t(\cdot,\cdot,\cdot)$. Upon learning each task, the model moves forward to learn the next task. Since all the previously learned tasks can be encountered at any time during testing in the future, the model should learn new tasks such that its knowledge of previously learned tasks is maintained, i.e., by preventing catastrophic forgetting. 

We formulate our problem in a \textbf{domain-incremental} learning setting~\cite{van2019three} which assumes all the tasks are from different domains and the boundaries between them are known during learning time. We consider that each task can be learnt individually by adapting a pre-trained large multimodal transformer  $f^{i}_{\theta M}(\cdot,\cdot)$ via minimizing a suitable  discrimination loss $\mathcal{L}$, e.g., cross entropy. In our approach,  all the model parameters, except the final classifier layer $\theta_{cls}$, are frozen during training to preserve the generalizability of the model. We benefit from prompt learning to enable using a single model to learn all   tasks. To prevent catastrophic forgetting, a trainable task-specific prompt pool, which contains several task-specific prompts, is attached to the model $f^{i}_{\theta M}(\cdot,\cdot)$ such that the best-semantically-matched prompt is selected based on   image and text inputs for task specialization. The prompt is then pre-pended to the input vectors so that the output is generated based on specialization. Our method is rehearsal-free and does not need any memory buffer similar to prior approaches \cite{lopezpaz2022gradient,rostami2023cognitively}.

\section{Proposed Architecture}
Our architecture, named cluster-based modality fusion prompt (\textbf{CluMo}), contains two unimodal task-specific cluster-based keys for vision and text embeddings and one prompt pool. The combination of the selections from both keys is then used to select the best matched prompt from the prompt pool. A high-level diagram of our approach is presented in Figure \ref{fig:Big_Diagram1}. In this section, we first introduce the preliminaries such as backbone model and prompt pool-based method in \ref{4.1}, and modality fusion prompt in Sec. \ref{4.2}, then the cluster-based prompt key is described in Sec. \ref{4.3}, and the training and the inference strategy is discussed in Sec. \ref{4.4}. 

\subsection{Preliminary}
\label{4.1}
\paragraph{Backbone}The base multimodal transformer contains three encoders: the visual encoder $\textit{VE}$, the textual encoder $\textit{TE}$, and the multimodal fusion encoder $\textit{FE}$. Given a visual input $\textbf{V}$, i.e., a single image, and a textual input $\textbf{T}$, i.e., a question, the data processing pipeline for the model is:
\begin{equation}
    \hat{y}(\textbf{V},\textbf{T}) = \mathcal{F}(\textit{FE}([\textit{VE}(\textbf{V});\textit{TE}(\textbf{T})])),
\end{equation}
where $\mathcal{F}(\cdot)$ is the classifier to predict the answer.

\paragraph{Prompt Pool} As an adoption of prompt learning in continual learning, a prompt pool is a set of trainable key-value ($K$-$P$) pair, in which $K \in R^{1\times D}$ denotes the ``prompt key'', and $P \in R^{L_p\times D}$ is the prompt. $L_p$ and $D$ denote the length and dimension of the prompt. Given an input image $\textbf{V}$, we compute $v_I=\textit{VE}(\textbf{V}) \in R^{L_v\times D}$, where $L_v$ is the dimension of the features,   after passing the image through the visual encoder. $v_{I_0} = v_I[0]$ is matched with all the keys $K$ within the prompt pool via similarity score, such as cosine similarity, to find the most similar $K_i$. The corresponding $P_i$ is selected and prepend to $\textbf{V}$ as $\textbf{V}'=[P_i;\textbf{V}]$. Parameters of $K$ and $P$ are updated through back-propagation during the training. However, in our setting, we adopt a two stage training strategy that prompt keys are trained before model and prompts.

\subsection{Modality Fusion Prompt}
\label{4.2}

Previous prompt-based CL methods such as L2P \cite{wang2022learning} associate each prompt in the prompt pool with a single prompt key to form Key-Value pair. In practice, the prompt keys in prompt-based CL can be considered as cluster centers.
These cluster encode a notion of similarity between the prompts. The input feature vectors that form a cluster in the feature space can be assigned to these cluster centers, which are prompt keys. The intuition behind this idea is that \textbf{feature vectors with small geometric distance   in the feature space are semantically similar} \cite{wang2023sprompts}. 

However, such a key-value pair design considers only single modality without tasks with multimodal inputs. The reason is that different input modalities contain different or complementary semantic information. Hence, having prompt keys that associate with each modality can help guiding prompt selection, which is more comprehensive and representative in term of semantic properties of each modality. Thus, we propose a task-specific prompt pool architecture, namely $\textbf{Modality Fusion Prompt}$,  which is composed of the \textbf{visual prompt keys} $K_v$, the \textbf{textual prompt keys} $K_t$, and the \textbf{prompt pool} $P$ as following: 
\begin{equation}
\begin{split}
      & K_t=[K_{t_1},K_{t_2},...,K_{t_{S_t}}], \\&
    K_v=[K_{v_1},K_{v_2},...,K_{v_{S_v}}], \\&
    P=[P_1,P_2...,P_{S_p}], \\&
    K_{t_m} \in R^{D}, K_{v_n} \in R^{D}, P_l \in R^{L_p \times D},
\end{split}
\end{equation}
 where $S_t$ , $S_v$, and $S_p$ are the sizes of textual prompt key,   the visual prompt key,  and the prompt pool, respectively. $L_p$ is the length of each prompt and $D$ is the hidden dimension of the transformer backbone. 
 The prompt pool size $S_p$ is then determined as $S_p = S_v \times S_t$. Each prompt is associated with the unique combination of one visual prompt key and one textual prompt key. Given a specific visual prompt key $K_{v_m}$ and a specific textual prompt key $K_{t_n}$, the Key-Key-Value pair is defined as the following:
\begin{equation}
    (K_{v_m}, K_{t_n}) \rightarrow P_{m * S_v + n}.
\end{equation}
As modality fusion prompt is \textbf{task-specific}, new visual prompt keys, textual prompt keys and a prompt pool will be initialized for each of the new coming task. The previous ones are frozen during training.

\subsection{Cluster-based Prompt Key}
\label{4.3}
\begin{algorithm}
\caption{PromptKeyTraining}
\label{Algorithm1}
\begin{algorithmic}
\REQUIRE  Dataset $D$, Image Prompt Key Pool $K_v$, Text Prompt Key Pool $K_t$, Image Prompt Size $S_I$, Text Prompt Size $S_T$  \\
\WHILE{Not Converge}
\STATE Random Select batch of image $I$, text $T$ from $D$
\STATE $\hat{v}_I = mean(\textit{VE}(I),dim=1)$
\STATE $\hat{v}_T = mean(\textit{VT}(T), dim=1)$
\STATE \textit{$Cluster_I$} = dictionary() \\
\STATE \textit{$Cluster_T$} = dictionary() \\
\FOR{\textit{i}, \textit{t} in $\textit{v}_{I_{mean}}$, $\textit{v}_{T_{mean}}$}
\STATE $Key_{img}$ = image key with top $similarity(i,K_v) $ \\
\STATE $Key_{txt}$ = text key with top $similarity(t,K_t)$ 
\STATE$Cluster_I$[$Key_{img}$].append(\textit{i})
\STATE $Cluster_T$[$Key_{txt}$].append(\textit{t})
\ENDFOR
\FOR{i in $\textit{S}_I$}
\STATE $\textit{K}_v$[i] = $mean$($Cluster_I$[i])
\ENDFOR
\FOR{i in $\textit{S}_T$}
\STATE $\textit{K}_t$[i] = $mean$($Cluster_T$[i])
\ENDFOR
\ENDWHILE
\end{algorithmic}
\end{algorithm}

$\textit{K}$-means clustering has been widely adopted in machine learning algorithms for semantic separation and understanding, where data from different domains can be explicitly separated via clustering in an unsupervised way \cite{wang2023sprompts} \cite{cohn2021unsupervised}. However, even though the data from single task belong to the same domain, they can still be further divided into sub-domains based on the semantic property. To make each prompt key be the semantically cluster center of the sub-domains for both vision and text inputs, we adopt mini-batch $K$-means clustering algorithm on prompt keys of $K_v$ and $K_t$ to make each prompt key diverse and representative. Let $\mathcal{B}$ = $(I, T)$ be the random batch from the training dataset. We extract the image feature vector $v_{I}$ and the text feature vector $v_{T}$ as follows:
\begin{equation}
    v_I = VE(I), v_T = TE(T),
\end{equation}
where $v_I \in R^{B\times L_I\times D}$ and $v_T \in R^{B\times L_T\times D}$, $B$ is the batch size, $L_I$ and $L_T$ are the length of vectors for image and text features, represent the embedded image and text input respectively. For visual prompt key clustering, each image feature vector, $v_{I_n}$, is set by taking mean along second the dimension such that $\hat{v}_{I_n} \in R^{B\times D}$, and $\hat{v}_{I_n}$ is used to compare with every prompt key in $K_v$:
\begin{equation}
    similarity(n,m) = ||\hat{v}_{{I_n}} - K_{v_m}||_2,
\end{equation}
and the prompt key with highest similarity is assigned to match $v_{I_n}$. After calculation of the whole batch $\mathcal{B}$, the prompt keys are then updated by calculating the mean of all $\hat{v}_{I_n}$ assigned to that specific visual prompt key. We repeat the above step until the convergence. The procedure of updating the text prompt key $K_t$ is similar to updating the image prompt keys. Algorithm \ref{Algorithm1} summarizes our approach for prompt key training.

\begin{figure*}[]
\centering
\includegraphics[scale=0.54]{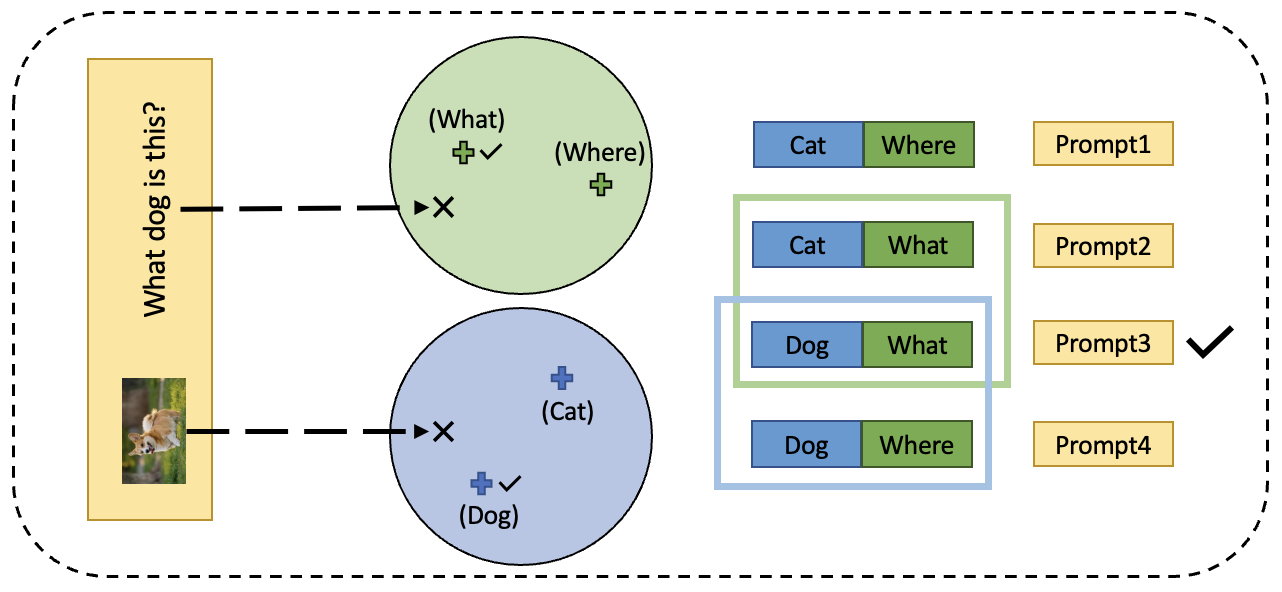}
\caption{Naive Example of Prompt Selection. Consider a naive animal VQA dataset which only contains dog and cat images and questions only about "what" and "where". During the first stage training, the visual prompt keys and textual prompt keys are learnt to represent "dog"/"cat" and "where"/"what" respectively (in realistic settings the keys will learn the implicit cluster instead of explicit category). Given a test image-question pair, the image and question are projected to their modality-specific feature space through the encoders. The nearest prompt keys, which are keys represent "dog" and "what" are selected. The combination of the two prompt keys lead to "prompt3" which is finally selected.}
\label{fig:Inference_Diagram}
\end{figure*}

\subsection{Training and Inference}

\begin{algorithm}
\small
\caption{Continual Learning Procedure}
\label{Algorithm2}
\begin{algorithmic}
\REQUIRE Datasets $\textbf{D}$, Pretrained Model $\textbf{M}$  \\
\STATE Freeze $\textbf{M}$ except $\textbf{M}$.classifier
\STATE $\textbf{M}$.CluMoList = dict() \\
\FOR{dataset $D_i$ in $\textbf{D}$}
\STATE Initialize new CluMo $\textbf{C}_i$ 
\STATE $\textbf{M}$.CluMoList[$D_i$] = $\textbf{C}_i$ \\
\STATE $\textbf{PromptKeyTraining}$($D_i$,$\textbf{C}_i$)\\
\STATE Freeze visual key and textual key of $\textbf{C}_i$
\FOR{Batch $\textbf{B}$ in $D_i$}
\FOR{img $\textit{i}$,txt $\textit{t}$ in $\textbf{B}$}
\STATE Select $Key_{img}$ based on $i$ from $\textbf{C}_i$
\STATE Select $Key_{txt}$ based on $t$ from $\textbf{C}_i$
\STATE Find prompt $P$ by $Key_{img}$ and $Key_{txt}$ from $\textbf{C}_i$
\ENDFOR
\STATE loss = \textbf{Train}($\textbf{M}$,$\textbf{B}$,$P$)
\STATE loss.backword()
\ENDFOR
\ENDFOR
\end{algorithmic}
\end{algorithm}
\label{4.4}
During training, we adopt a $\textbf{two-stage training}$ strategy to ensure that the prompt keys are correctly settled before learning the current task. In the first stage of learning each task $T_i$, we random select batches from the current task's dataloader to train minibatch $k$-means Cluster on the visual and the textual prompt key $K_v$ and $K_t$ until reaching the convergence of the clustering algorithm. During the second stage, the trained $K_v$ and $K_t$ are frozen. Within the iteration of training dataloader, each training instance is assigned to its nearest prompt key using $k$-nearest neighbor (KNN) algorithm to find the best match prompt $P_k$ from the prompt pool $P$. $P_k$ is then attached to the model pipeline:
\begin{equation}
\hat{y}(\textbf{V},\textbf{T}) = \mathcal{F}(\textit{FE}([P_k;\textit{VE}(\textbf{V});\textit{TE}(\textbf{T})]))
\end{equation}

During the second stage, we also use knowledge distillation to further boost the performance. Before the training of task $T$, we keep a frozen copy of model after finishing $T-1$, denoted as $\mathcal{M}_{T-1}$. To prevent   significant parameter shift, we pass the same input image and question to both $\mathcal{M}_T$ and $\mathcal{M}_{T-1}$ and add the difference between the two model's output to the loss:
\begin{equation}
    \mathcal{L}_{KD}(\textbf{V},\textbf{T}) = MSE(\hat{y}_{\mathcal{M}_T}(\textbf{V},\textbf{T}), \hat{y}_{\mathcal{M}_{T-1}}(\textbf{V},\textbf{T})).
\end{equation}
The final  objective loss function would be:
\begin{equation}
    \mathcal{L} = \mathcal{L}_{ce}(\hat{y}(\textbf{V},\textbf{T}),y) + \mathcal{L}_{KD}
\end{equation}
Where $\mathcal{L}_{ce}$ is the same cross entropy loss. The overall training procedure for all tasks come in sequence is presented in Algorithm \ref{Algorithm2}.

During   inference,  the   model is frozen and we follow a  procedure similar to the  second stage  of training. For every training image-text pair, the image input is aligned with the best-matched image prompt key while the text input is aligned with the best-matched text prompt key. The combination of prompt keys is deployed to find the corresponding prompt, which is pre-pend to the output of multimodal encoder. To help better understand the procedure of prompt selection, an visualization example is provided in Figure \ref{fig:Inference_Diagram}.

\section{Experiments}

Our implementation code is available at \url{https://github.com/YuliangCai2022/CLUMO.git}. Please refer to the code for reporducing the results.

\subsection{Experiment Setup}

\paragraph{Backbone} 
We used the public pre-trained large multimodal transformer, ALBEF \cite{li2021align}, as our backbone for VQA task. It consists of an image encoder, a text encoder, a multimodal encoder, which uses cross-attention between the two modalities. Specifically for VQA tasks, an pre-trained answer decoder is append after the multimodal encoder, which has same architecture as multimodal encoder.

\paragraph{Baselines for comparison} 
We use seven methods for comparison. We include algorithms from major CL approaches. We include two regularization-based methods: $\textbf{EWC}$ \cite{Kirkpatrick_2017} and $\textbf{LwF}$ \cite{li2017learning}, two rehearsal-based methods: $\textbf{ER}$ \cite{rolnick2019experience} and $\textbf{GEM}$ \cite{su2021gem}. We also include three SOTA prompt-based continual learning methods, $\textbf{L2P}$ \cite{wang2022learning}, $\textbf{DualPrompt}$ \cite{wang2022dualprompt}, and $\textbf{S-Prompt}$ \cite{wang2023sprompts}. We also include finetuning to demonstrate the positive effect of CL. Following the original setting of each method, we leave the whole backbone model unfrozen for non-prompt-based methods and freeze the whole backbone model for prompt-based methods except for the classifier. To make the fair comparison, we fit all the continual learning methods into our backbone, ALBEF, instead of using the original model proposed in each method.

\paragraph{CL Tasks}
We evaluate our method on tasks built using the $\textbf{CLOVE}$ \cite{lei2022symbolic} dataset which is a VQA-based continual learning dataset. The benchmark contains two sepecate benchmarks for different scenario, including scene-incremental setting benchmark, $\textbf{CLOVE-scene}$, and function-incremental setting benchmark, $\textbf{CLOVE-function}$. Each of the task sets contains six tasks which are domain-specific and diverse from each other. For more details about   $\textbf{CLOVE}$
and the tasks we use, please refer to  the Appendix. 

\paragraph{Metrics for comparison} We use the average accuracy and the average forgetting rate on all tasks to evaluate the performance of our method and its ability to tackle catastrophic forgetting. Different from dataset such as \textbf{VQAv2} \cite{goyal2017making}, where each question is paired with different ground truth answers, questions in \textbf{GLOVE} dataset only contains exactly one correct answer for each question. Thus, the accuracy is simply calculated by $y==\hat{y}$ for every training and testing data instance. On the other hand, the forgetting rate is calculated as:
\begin{equation}
    F = \frac{A_i-A_{ij}}{A_i}
\end{equation}
where $A_i$ is the accuracy of task $i$, and $A_{ij}$ is the accuracy of task $i$ after the model is trained on task $j$. For details about the optimization and implementation processes, please refer to the Appendix.


\subsection{Comparative Results}

\begin{table*}[]
\small
\centering
\scalebox{0.88}{
\begin{tabular}{c||c|c||c|c||c|c||c|c||c|c||c|c}
    \toprule
    & \multicolumn{12}{c}{CLOVE-scene}\\
    \textbf{Method}& \multicolumn{2}{c||}{abcdef} & \multicolumn{2}{c||}{dbafec} & \multicolumn{2}{c||}{bdcafe} & \multicolumn{2}{c||}{acbefd} & \multicolumn{2}{c||}{caefdb} & \multicolumn{2}{c}{bafedc}\\     
     & A $\uparrow$ & F $\downarrow$& A $\uparrow$& F $\downarrow$& A $\uparrow$ & F 
     $\downarrow$& A  $\uparrow$ & F $\downarrow$& A $\uparrow$ & F $\downarrow$& A $\uparrow$ & F $\downarrow$\\ \midrule
    Finetune & 34.03 & 34.28 & 34.89 & 34.99 & 38.83 & 21.65 & 34.45 & 35.14 & 34.42 & 34.47 & 33.95 & 35.67 \\ 
    \midrule
    EWC & 37.49 & 28.04  & 37.00 & 29.10 & 37.95 & 27.46 & 37.99 & 29.68 & 37.13 & 28.63 & 37.83 & 27.97\\ 
    LwF & 38.18 & 26.82 & 35.03 & 32.84 & 37.31 & 29.11 & 37.85 & 29.87 & 37.94 & 28.15 & 38.21 & 27.48\\ 
    \midrule
    ER & 41.05 & 19.92 & 42.09 & 17.12 & 42.37 & 18.09 & 41.91 & 20.28 &  41.11 & 19.65 & 42.08 & 20.52\\ 
    GEM & 41.52 & 18.33 & 43.14  & 14.73 & 42.89 & 17.43 & 42.54 & 20.13 & 41.90 & 20.88 & 43.11 & 19.86  \\ 
    \midrule
    L2P & 43.01& 18.22& 45.84 & 15.03 & 44.64 & 17.41 & 45.63 & 14.96 & 44.78 & 17.99 & 46.58 & 14.85\\ 
    DualPrompt & 45.51 & 15.86 & 46.58 & 13.49 & 45.83 & 16.48 & 46.27 & 15.45 & 46.21 & 15.89 & 47.01 & 13.16\\ 
    S-Prompt & 45.73 & 14.11& 45.93 & 14.17& 46.99 & 14.38 & 46.68 & 14.77 & 47.53 & 12.19 & 44.09 & 22.32 \\ \midrule
    \textbf{CluMo} & \textbf{48.73} & \textbf{10.76} & \textbf{48.8} & \textbf{10.25} & \textbf{48.83} & \textbf{9.73}  & \textbf{48.94} & \textbf{10.29} & \textbf{48.26} & \textbf{11.04} & \textbf{48.98} & \textbf{10.23} \\
    \bottomrule
    \toprule
    & \multicolumn{12}{c}{CLOVE-function}\\
    \textbf{Method}& \multicolumn{2}{c||}{soarkl} & \multicolumn{2}{c||}{rsaolk} & \multicolumn{2}{c||}{osrlak} & \multicolumn{2}{c}{oarlks} & \multicolumn{2}{c}{skaolr} & \multicolumn{2}{c}{ksoarl} \\     
     & A $\uparrow$ & F $\downarrow$& A $\uparrow$& F $\downarrow$& A $\uparrow$ & F 
     $\downarrow$& A  $\uparrow$ & F $\downarrow$& A $\uparrow$ & F $\downarrow$& A $\uparrow$ & F $\downarrow$\\ \midrule
    Finetune & 31.55 & 53.76  & 37.34 & 39.64 & 23.34 & 57.32 & 24.09 & 62.79& 16.34 & 74.82 & 17.50 & 84.46 \\ 
    \midrule
    EWC & 35.70 &  47.92 & 37.82 & 41.55 & 38.92 & 40.48 & 40.74 & 33.89 & 37.53 & 37.22 & 40.85 & 32.32\\ 
    LwF & 37.18 & 46.86 & 36.81 & 44.12 & 39.21 & 39.81 & 36.81 & 41.29 & 30.49 &  53.11 & 29.17 & 55.84\\ 
    \midrule
    ER & 42.22 & 32.97 & 39.78 & 38.62 & 41.22 & 35.79 & 37.14 & 33.38 & 33.41 & 48.99 & 38.23 & 38.01\\ 
    GEM & 44.58 & 30.87 & 41.43 & 29.46 & 40.87 & 32.98 & 39.81 & 28.77& 36.88 & 39.14 & 40.26 & 31.87  \\ 
    \midrule
    L2P & 44.80 & 16.38 & 43.39 & 21.26 & 43.27 & 21.97 & 42.54 & 19.18 & 40.4 & 31.92 & 43.37 & 24.19\\ 
    DualPrompt & 45.01 & 15.90 & 44.26 &  17.43 & 44.66 & 18.50 & 43.69 & 15.31  & 39.32& 34.78 & 45.65 & 20.54\\ 
    S-Prompt & 45.45 & 13.47 & 45.01 & 14.76 & 45.27 & 14.29 & 42.98 & 20.20 & 42.85 & 25.82 & 44.09 & 22.32 \\ \midrule
    \textbf{CluMo} & \textbf{46.18} & \textbf{10.62} & \textbf{45.36} & \textbf{11.69} &  \textbf{46.34} & \textbf{10.22} & \textbf{45.95} & \textbf{9.15} & \textbf{45.66} & \textbf{19.89} & \textbf{46.89} & \textbf{17.41} \\
    \bottomrule
\end{tabular}
}
\caption{Comparative experimental results: the accuracy and forgetting rate for different task order. For each task sequence, \textbf{A} $\uparrow$ indicates the accuracy of the method, while \textbf{F} $\downarrow$ is the forgetting rate of each.}
\label{tab:comparative}
\end{table*}

We conduct the comparison experiments on both the $\textbf{CLOVE-scene}$ and $\textbf{CLOVE-function}$ task sets with a randomly selected task order. In table \ref{tab:comparative}, the task order $\textit{abcdef}$ represents the CL tasks: $\textit{ShopAndDining}$, $\textit{WorkPlace}$, $\textit{HomeOrHotel}$, $ \textit{Transportation}$, $\textit{SportAndLeisure}$  $\textit{Outdoors}$ in sequence. The $\textit{oarlks}$ in $\textbf{CLOVE-function}$ represents tasks: $\textit{ObjectRecognition}$, \textit{AttributeRecognition}, \textit{RelationReasoning}, \textit{LogicReasoning}, \textit{KnowledgeReasoning} and \textit{SceneTextRecognition}.

We observe in Table \ref{tab:comparative} that our method outperforms all the   baselines across all task order sets in terms of both average accuracy and average forgetting rates. We also observe that the performance of different method within the same group tend to be similar. The regularization-based methods, $\textbf{EWC}$ and $\textbf{LwF}$, obtain the sub-optimal accuracy and forgetting rate besides. The reason is that the domain for each task in the dataset is significantly different from the rest of tasks and hence regularization methods fail to capture the common space of the parameter distribution. This challenge makes it difficult to maintain the accuracy of the current task and previous tasks at the same time using regularization. The replay methods, \textbf{ER} and \textbf{GEM},   achieve better performance than regularization-based methods. This can be explained by the fact that replaying the data from previous task is an efficient way to remind the model and adjust its parameter distribution not too diverse from previous ones. However, because we need to rely on a memory buffer to store samples for replay, these methods are memory-consuming and thus not space-efficient. Moreover, replay-based methods are still limited by the upper-bound of joint training, as they generally can only reduce catastrophic forgetting without boosting the accuracy of individual tasks. On the other hand, the prompt-based methods, namely \textbf{L2P}, \textbf{DualPrompt}, and \textbf{SPrompt}, achieve superior performances compared to more traditional CL methods. Rather than tuning the whole model with regularization, prompt-based methods store the prior knowledge in trainable prompts, which are smaller and more efficient than memory buffer, and keep the main body of backbone model frozen. With the combination of generalization capacity of pre-trained model and specific previous knowledge stored in prompt, prompt-based method can outperform  the replay and regularization methods. Among all the methods, our method achieve the best performnce.

Compared with the baseline prompt-based methods which only consider visual modality for prompt selecting and updating, \textbf{CluMo} takes care of both the visual and textual modalities, as well as the fusion of the two for selecting  the prompt which deploys the given information more comprehensively to process the prompt. Our design thus fits better in multimodal learning scenario than other existing continual learning methods.

\subsection{Ablation Experiments}

\begin{table}
\centering
\caption{Ablative Experiments}
\begin{tabular}{c||c|c}
\toprule
Methods & Accuracy & Forgetting \\
\midrule
Full Method & 48.73 & 10.76 \\
\midrule
Ablative KD & 47.36 & 11.25\\
Ablative Clustering & 46.08 & 12.86\\
Ablative Textual Key & 46.16 & 12.49\\
Ablative Visual Key & 46.53 &  12.22\\
\bottomrule
\end{tabular}
\label{tab:ablative}
\end{table}

To offer a better insight about our method, we perform an ablation study for each component of \textbf{CluMo} to study the positive contribution of each component. We study  the effect of the following:
\begin{itemize}
    \item \textbf{Visual Prompt Key}, key to separate the inner-task image features by their semantic property.

    \item \textbf{Textual Prompt Key}, key to separate the inner-task text features by their semantic property.
    \item \textbf{Minibatch $k$-means Clustering} which train the prompt keys as centers of clustering algorithm to better fit the semantic meaning.
    \item \textbf{Knowledge Distillation}, to prevent the drastic parameter shift of unfrozen classifier.
\end{itemize}

We conduct  ablation experiment on \textbf{CLOVE-scene} dataset with the task order \textit{abcdef}. We set the size for both the visual prompt key and the textual prompt key to be three. For ablative text experiments, we change the size of textual prompt key to 9 to achieve the same prompt size. We also removed the visual prompt key which is the same for ablative image experiments. Results for this experiment is presented in Table \ref{tab:ablative}. We observe that despite having the same number of prompts, the performance values of Ablative Textual Key and Ablative Visual Key are lower than our full pipeline. This result verifies our hypothesis that both modalities should be used to guide the prompt selection and the missing of any will cause information lost and lead to sub-optimal performance. In other words, current approaches for unimodal settings do not use all the information we have in multimodal scenarios. We also observe that without the clustering algorithm, the performance of ablative clustering is the lowest among all the settings which indicate the significance of doing cluster training for learning the prompt keys.

\begin{figure}
\centering
\includegraphics[scale=0.45]{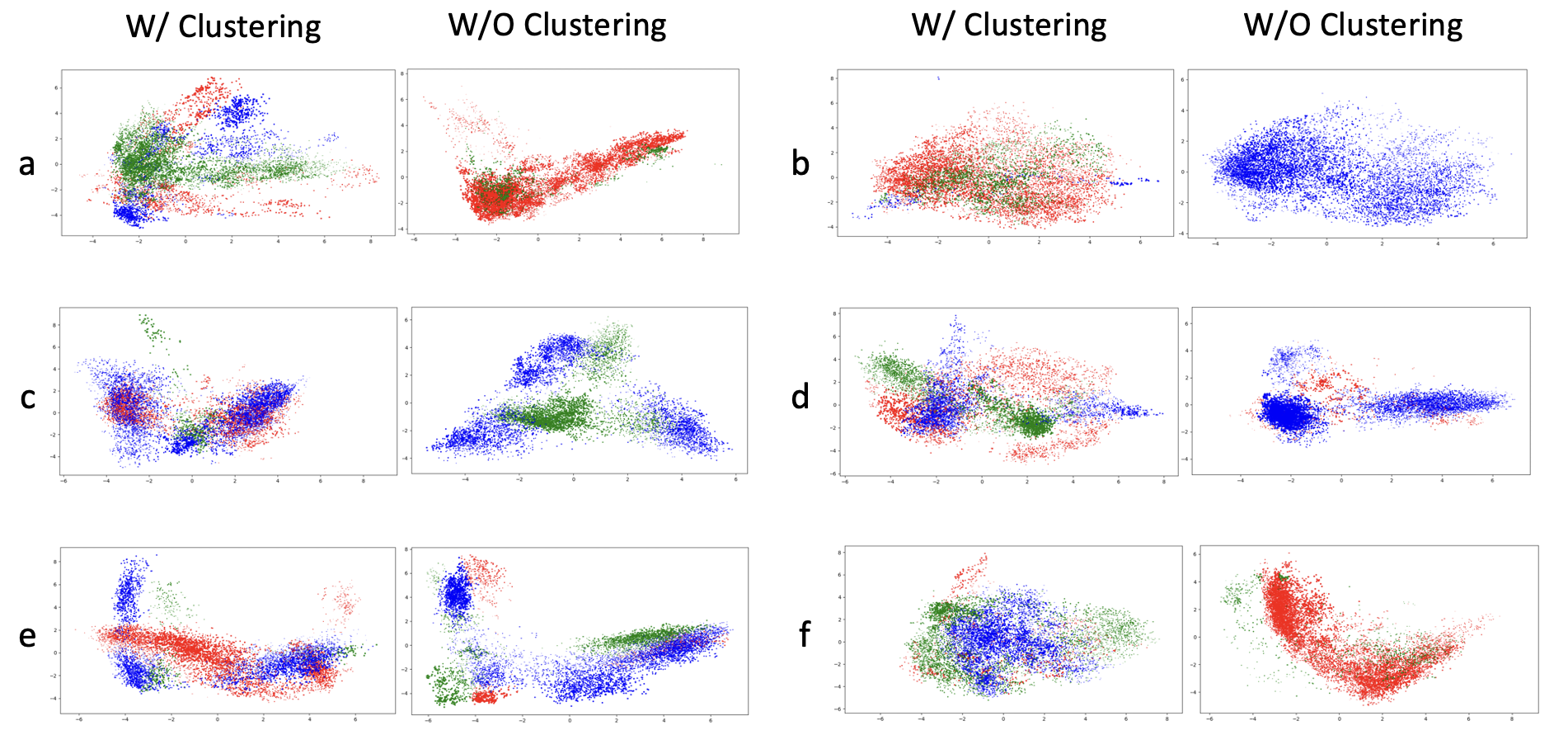}
\caption{Cluster distribution on all training image data of $\textbf{CLOVE-Scene}$'s six sub-tasks before and after applying mini-batch $k$-means clustering algorithm with visual key size of 3 using PCA. \textbf{Color of more diversity indicates more even distribution of key selection.}}
\label{fig:Cluster_visual}
\end{figure}

\begin{figure}
\centering
\includegraphics[scale=0.75]{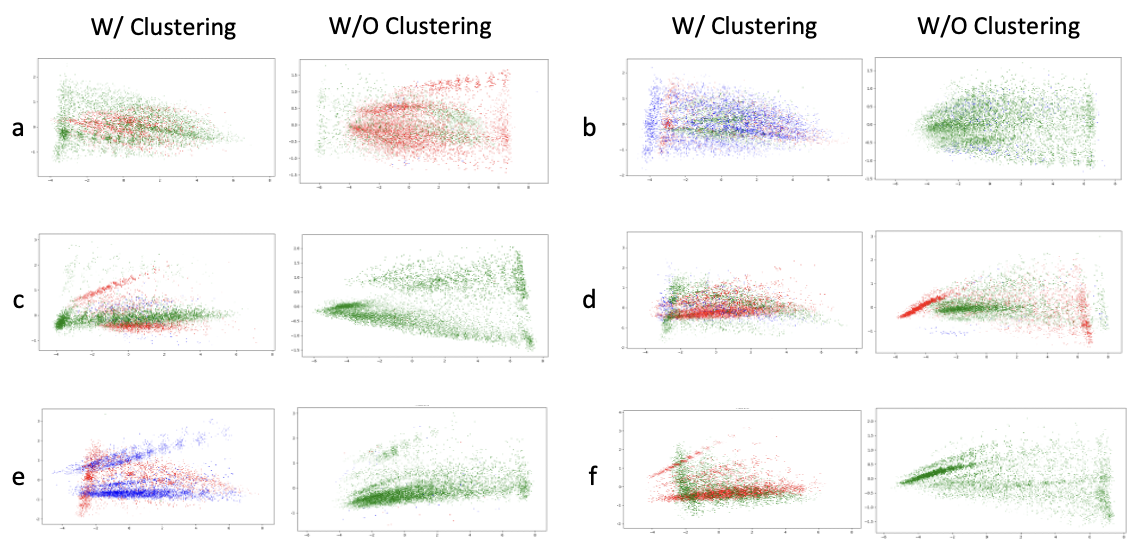}
\caption{Cluster distribution on all training text data of $\textbf{CLOVE-Scene}$'s six sub-tasks before and after applying mini-batch $k$-means clustering algorithm with textual key size of 3 using PCA. }
\label{fig:Cluster_textual}
\end{figure}

\subsection{Analytic Experiments}

\begin{figure}[]
\centering
\includegraphics[scale=0.5]{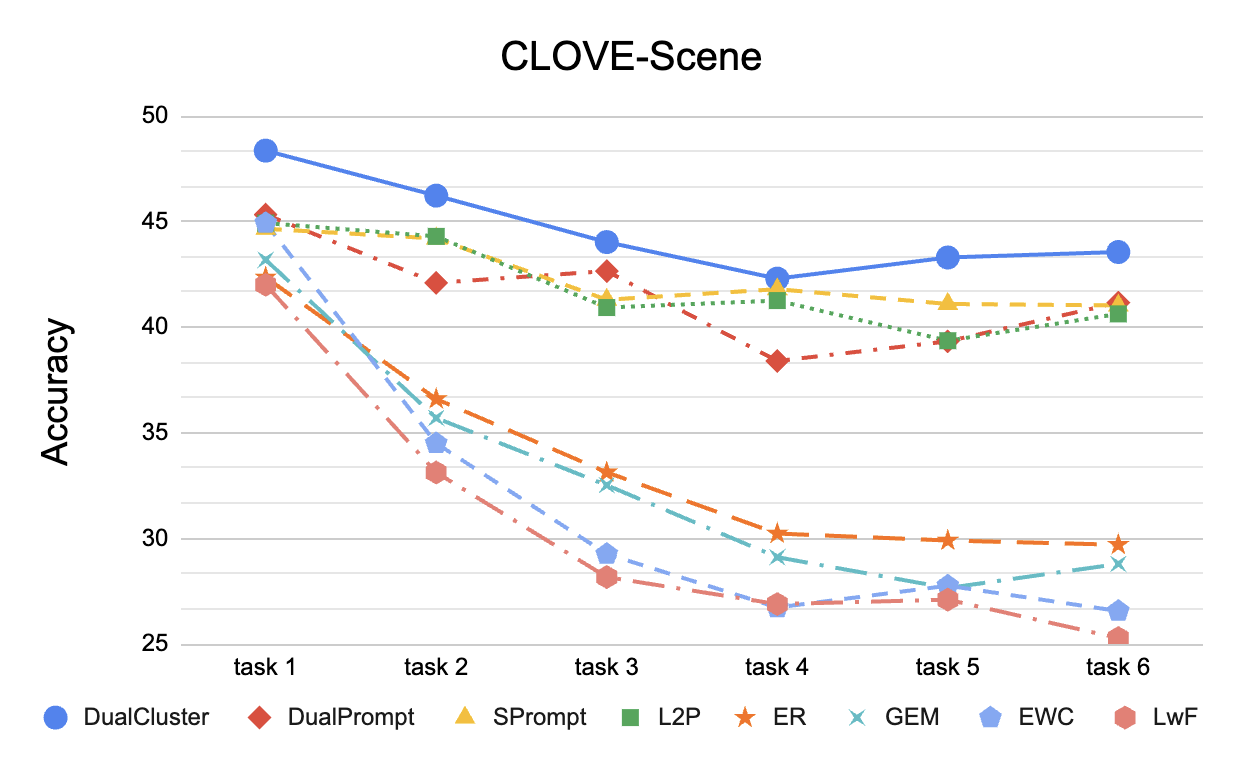}
\caption{Accuracy on the first task after running task sequence.}
\label{fig:First_exp_scene}
\end{figure}

\paragraph{Effect of clustering}
To show the effect of clustering algorithm, we empirically show the correlation between the clustering error and the downstream accuracy. As we apply Euclidean distance as metric to learn the clusters, we record the average distance between each point to its assigned cluster center for every task, and take the average for all the tasks:
\begin{equation}
    \mathcal{E} = Avg(\sum^N_{i=1}Avg(\sum^M_{j=1}||x_j-c_k||_2))
\end{equation}
where $i$ represent the number of tasks, $j$ represent the training data from task $i$ and $k$ is the $k^{th}$ cluster center. We  consider both the visual prompt key training and the textual prompt key training in this experiment. Table \ref{tab:clustering_error}  presents the results. Same as our expectation, we observe a negative correlation between the clustering error and the performance accuracy, in another word,  lower $\mathcal{E}$ for image and text prompt keys leads to a higher  accuracy. Without the  clustering component, we observe $\mathcal{E}$ to be as high as 42.38 and 42.8 for image prompt key and text prompt key, respectively. After applying clustering algorithm,  $\mathcal{E}$ drops below 20, which can be considered significant, for both modalities and the accuracy improves 2.97\%.

\paragraph{Cluster Visualization}
To show the effect of clustering on prompt key more intuitively, we visualize the visual prompt key selection distribution and textual prompt key selection distribution on the visual and textual portion of the training data for $\textbf{CLOVE-Scene}$ in Figure \ref{fig:Cluster_visual} and Figure \ref{fig:Cluster_textual}. Since we use three visual prompt keys for each task, the vector features of visual data are split into three groups, which are the green, blue and red points in Figure \ref{fig:Cluster_visual}. We observe that without using clustering, visual data are more likely to overlap on the same cluster center which means they would lead to select the same visual prompt key. After performing  clustering, we observe that the distribution becomes more evenly, and every cluster of data is diverse and separated from the others which means that the visual data can be separated explicitly. The diversity of prompt key selection indicates each input image can find the 
"correct" prompt which is semantically closer to it, which contains more specific knowledge about the sub-domain the given image belongs to. The visualization of textual prompt keys indicates similar observation.

\begin{table}[h]
\centering
\begin{minipage}{0.45\textwidth}
\centering
\caption{Acc. with different clustering error}
\begin{tabular}{c|c||c}
\toprule
$\mathcal{E}$. Image & $\mathcal{E}$. Text & Accuracy \\
\midrule
15.40 & 10.72 & 48.73 \\
15.74 & 12.22 & 48.03\\
 17.21 & 12.53 & 47.94\\
42.38 & 42.8 & 47.32\\
\bottomrule
\end{tabular}
\label{tab:clustering_error}
\end{minipage}
\hfill
\begin{minipage}{0.45\textwidth}
    \centering
\caption{Acc. with different prompt key size}
\begin{tabular}{c||c}
\toprule
 $S_{img} \times S_{txt}$ & Accuracy \\
\midrule
2 $\times$ 2 & 48.51 \\
3 $\times$ 3 & 48.73\\
4 $\times$ 4 & 48.32\\
5 $\times$ 5 & 48.32\\
10 $\times$ 10 & 48.51 \\
\bottomrule
\end{tabular}
\label{tab:size_compare}
\end{minipage}
\end{table}
\paragraph{Tracking the Accuracy for the First Task}
To take a closer look into the effect on preventing catastrophic forgetting and increasing the accuracy in CL, we track the accuracy of the first task while learning the task sequence. The result is shown in Figure \ref{fig:First_exp_scene}. We see that the accuracy drops until task 4, and then slightly increases until task 6. Although it is not our main focus, this behavior shows a trend of forward transfer between the tasks. 
Among all the baseline methods, we notice that prompt-based methods, \textbf{SPrompt}, \textbf{DualPrompt} and \textbf{L2P}, significantly outperform other methods which verifies the SOTA status of prompt learning in CL and its success in preventing catastrophic forgetting. Our method $\textbf{CluMo}$, on the other hand, still outperform all prompt-based baseline methods. We observe that using the cluster-based prompts, the accuracy on the first task is superior compared to the other methods at the very beginning. Similar to other prompt-based method, our method's accuracy slightly drops until task 4 and improves subsequently. As the accuracy of our proposed method is higher than others at all time steps, our method has the leading performance in terms of both accuracy and backward transfer.

\paragraph{Effect of Prompt Key Size}
We also conduct an experiment to study the effect of prompt pool size to show the stability of our method with respect to this hyper-parameter. In Table \ref{tab:size_compare}, we choose different visual prompt key and textual prompt key sizes, $2 \times 2, \ 3 \times 3, \ 4 \times 4, \ 5 \times 5, \ 10 \times 10 $, corresponding to 4, 9, 16, 25, and 100 prompt pool sizes. Although having variant sizes, we only observe minor changes in accuracy in Table \ref{tab:size_compare}  when  prompt pool size changes, i.e., between 48.32 and 48.73. This observation means that our method is not sensitive to the change of the prompt pool size and hence we don't need to focus on tuning it.

\paragraph{Prompt Length and Prompt Size}

In the experiment setting, we utilize 3 visual keys and 3 textual keys, and each prompt has length of 10. In sum, we have $3\times3\times10=90$ total prompt length for each task. We conduct experiments to show the trade-off of prompt length and prompt numbers given the fixed total prompt length. To make a fair comparison, we keep the visual key and textual key the same size, and choose the closest integer prompt length to let the multiplication of the three be around 90 . Within table \ref{tab:tradeoff}, among the three different combination, we find that the current setting, which is the most balanced one, obtains the best result. Regarding the rest of the two, the higher prompt length has the better performance. 
\begin{table}
\centering
\caption{Acc. with length/number trade-off}
\begin{tabular}{c||c}
\toprule
$S_{v} \times S_{t} \times L_p$ & Accuracy \\
\midrule
$2\times2\times22$ & 47.84 \\
$3\times3\times10$ & 48.73\\
$4\times4\times6$ & 46.59\\
\bottomrule
\end{tabular}
\label{tab:dataset}
\end{table}
\section{Conclusion}
We introduced a novel prompt-based continual learning method for learning multimodal tasks. While most of existing methods apply simple prompts on a single modality, our method proposes modal-specific visual prompt keys and textual prompt keys and train them to capture the semantic properties of the training dataset using K-means clustering algorithm. We use the combination of both the visual prompt key and the textual prompt key to select prompts, which enable the prompt to better boost the performance. Our experiments show that our method achieves the state-of-the-art performance in continual VQA tasks in different domains compared to other regularization-based, rehearsal-based and prompt-based CL methods. 


\bibliography{main}
\bibliographystyle{tmlr}

\clearpage

\appendix
\section{Appendix}
\subsection{Hardware Setup and Hyper-parameter}

All experiments were conducted using a single Nvidia A40 GPU. We employed the AdamW optimizer across all experiments, utilizing a cosine learning rate scheduler, and set the initial learning rate to $lr = 3 \times 10^{-4}$. The models were trained for 5 epochs with a batch size of 16.

For $\textbf{EWC}$, we set the fisher sample percentage to be 0.1 and ewc loss weight equals to 0.1 as well.

For $\textbf{Experiment Replay}$, we store 1\% of data from each tasks into the memory buffer. During the training of the current task, we randomly select a batch of data from the memory buffer to train the model for every 100 batches of current data training.

For $\textbf{GEM}$, we also store 1\% of data to memory buffer, which is randomly picked from each tasks.

For $\textbf{CluMo}$ framework, we configured the visual prompt key size ($S_v$) and the text prompt key size ($S_t$) both to 3, with the prompt length ($L_p$) set to 10.

For $\textbf{L2P}$, we set the prompt pool size equals to 20 and prompt length to be 5.

For $\textbf{DuamPrompt}$, the G-Prompt was inserted into layers 0 and 1 of the visual encoder, while the E-Prompt was integrated into layers 2, 3, and 4. 

For $\textbf{S-Prompt}$, we set the prompt length to be 10 and prompt pool size to be 30.

Furthermore, for prompt-based techniques such as L2P, DualPrompt, and SPrompt, we opted to freeze the entire backbone model, allowing only the final classifier layer to remain trainable. Conversely, for all other baseline methods, no parameters were frozen, ensuring the entire network was fine-tuned during training.

\subsection{CLOVE dataset detail description}

In the \textbf{CLOVE-Scene} and \textbf{CLOVE-Function} datasets, all tasks have a uniform distribution of training and testing data, with the exception of the $SceneTextRecognition$ task, which comprises 16.8K training samples and 2.4K testing samples. The remaining tasks within these datasets contain 20K training samples and 3K testing samples each. It is reflected in the Table \ref{tab:dataset}.

\begin{table}
\centering
\caption{size of each task in CLOVE}
\begin{tabular}{c|c|c|c}
\toprule
task & training size & testing size & image source \\
\midrule
ShopAndDining & 20K & 3K & MS-COCO\\
WorkPlace & 20K & 3K & MS-COCO\\
HomeOrHotel & 20K & 3K & MS-COCO\\
Transportation & 20K & 3K & MS-COCO\\
SportAndLeisure & 20K & 3K & MS-COCO\\
Outdoors & 20K & 3K & MS-COCO\\
ObjectRecognition & 20K & 3K & MS-COCO\\
AttributeRecognition & 20K & 3K & MS-COCO\\
RelationReasoning & 20K & 3K & MS-COCO\\
LogicReasoning & 20K & 3K & MS-COCO \\
KnowledgeReasoning & 20K & 3K & MS-COCO\\
SceneTextRecognition & 16.8K & 2.4K & VG\\
\bottomrule
\end{tabular}
\label{tab:tradeoff}
\end{table}

\begin{figure*}[ht]
\centering
\includegraphics[scale=0.74]{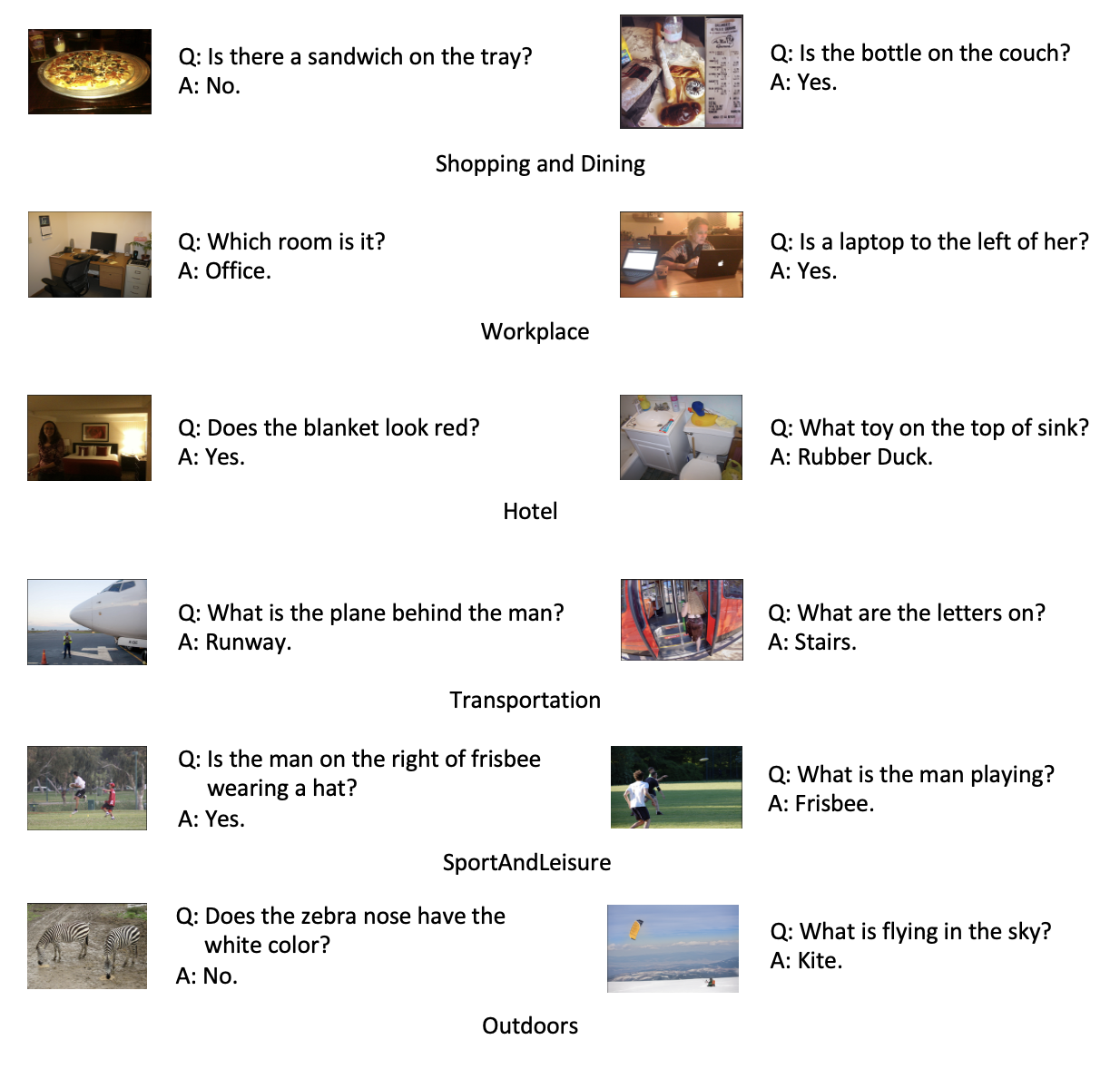}
\caption{CLOVE-scene dataset samples}
\label{fig:CLOVE_scene}
\end{figure*}

\begin{figure*}[ht]
\centering
\includegraphics[scale=0.74]{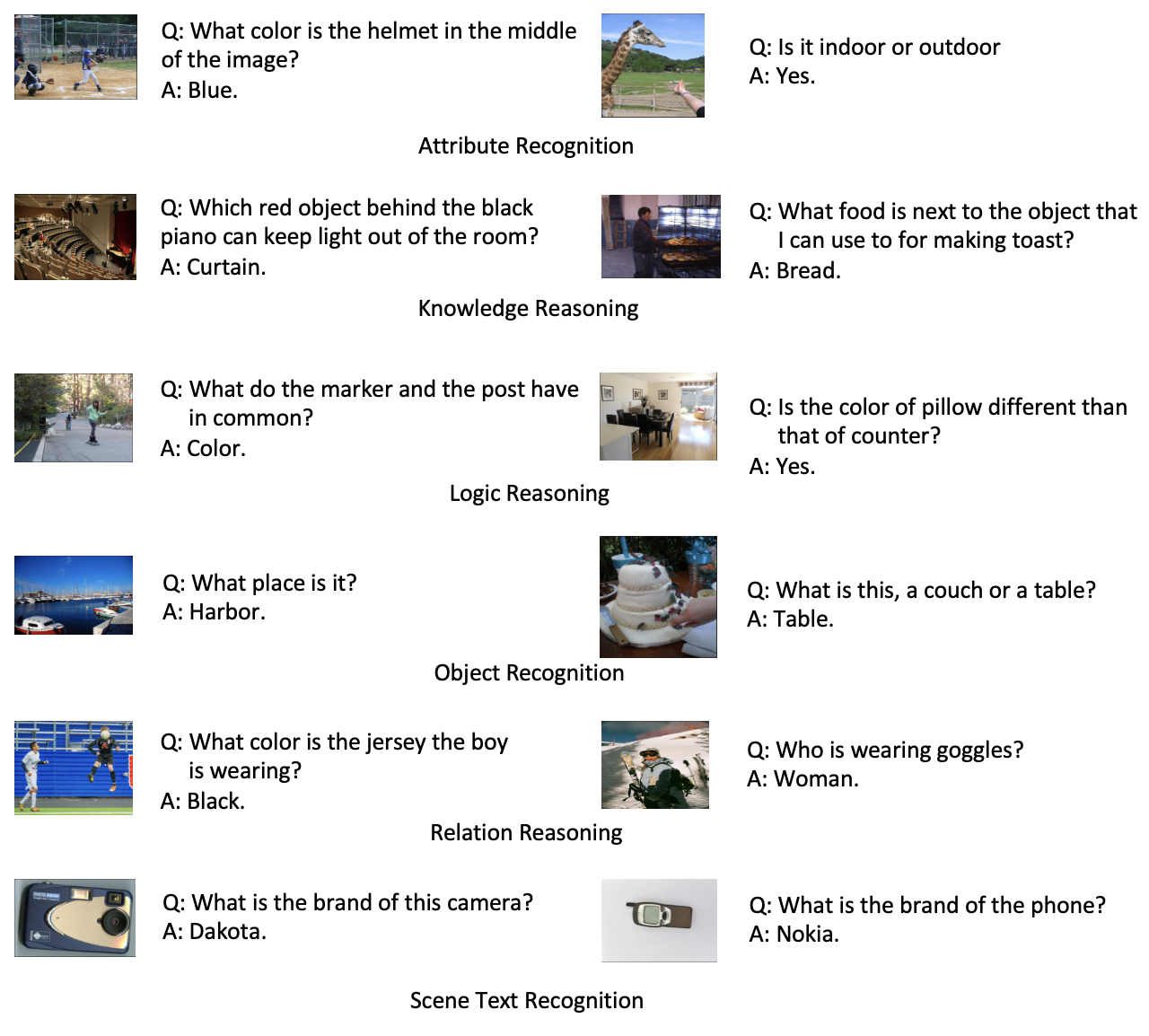}
\caption{CLOVE-function dataset samples}
\label{fig:CLOVE_function}
\end{figure*}

To provide a deeper understanding of the $\textbf{CLOVE}$ dataset, we offer additional details here. We have included visualizations in Figure \ref{fig:CLOVE_scene} and Figure \ref{fig:CLOVE_function} to showcase two sample images from each task within the datasets, emphasizing the distinctiveness of each domain. From these samples, it is evident that the images in the \textbf{CLOVE-Scene} dataset vary significantly across tasks, even though the questions associated with them are similar in structure, differing primarily based on the content depicted in the images.

On the other hand, the \textbf{CLOVE-Function} dataset presents a different scenario. While the images across various tasks may appear to belong to similar or overlapping domains, making it challenging to distinguish one task from another based solely on visual content, the diversity becomes apparent when considering the questions. Each task within the \textbf{CLOVE-Function} dataset involves distinct types of reasoning, as reflected in the varied nature of the questions posed, which are tailored to serve different reasoning purposes.

\end{document}